\documentclass[10pt,twocolumn,letterpaper]{article}

\usepackage{cvpr}
\usepackage{times}
\usepackage{epsfig}
\usepackage{graphicx}
\usepackage{amsmath}
\usepackage{amssymb}

\usepackage{enumitem}
\usepackage{subcaption}
\usepackage{booktabs}
\usepackage{multirow}
\usepackage{makecell}
\usepackage[nocompress]{cite}
\usepackage{soul}

\usepackage[pagebackref=true,breaklinks=true,letterpaper=true,colorlinks,bookmarks=false]{hyperref}

\usepackage{mmstyle}

\cvprfinalcopy % *** Uncomment this line for the final submission

\def\cvprPaperID{633} % *** Enter the CVPR Paper ID here

\ifcvprfinal\fi
\begin{document}

%%%%%%%%% TITLE
\title{Optimizing Video Object Detection via a Scale-Time Lattice}

\author{Kai Chen$^1$ Jiaqi Wang$^1$ Shuo Yang$^{1,2}$ Xingcheng Zhang$^1$
Yuanjun Xiong$^{1,2}$ Chen Change Loy$^1$ Dahua Lin$^1$\\
$^1$CUHK - SenseTime Joint Lab, The Chinese University of Hong Kong \hspace{10pt} $^2$Amazon Rekognition\\
{\tt\small \{ck015,wj017,zx016,ccloy,dhlin\}@ie.cuhk.edu.hk}\hspace{1cm}
{\tt\small \{shuoy,yuanjx\}@amazon.com}
}

\maketitle
\thispagestyle{empty}

%%%%%%%%% ABSTRACT
\begin{abstract}
    % !TEX root = ../main.tex

High-performance object detection relies on expensive convolutional
networks to compute features, often leading to significant challenges in
applications, \eg~those that require detecting objects from
video streams in real time.
The key to this problem is to trade accuracy for efficiency in an effective
way, \ie~reducing the computing cost while maintaining
competitive performance.
To seek a good balance, previous efforts usually focus on optimizing the
model architectures. This paper explores an alternative approach, that is,
to reallocate the computation over a scale-time space.
The basic idea is to perform expensive detection sparsely and
propagate the results across both scales and time with
substantially cheaper networks, by exploiting the strong correlations
among them.
Specifically, we present a unified framework that integrates detection,
temporal propagation, and across-scale refinement on a Scale-Time Lattice.
On this framework, one can explore various strategies to balance performance and cost.
Taking advantage of this flexibility, we further develop an adaptive scheme
with the detector invoked on demand and thus obtain improved tradeoff.
On ImageNet VID dataset, the proposed method can achieve a competitive mAP
$79.6\%$ at $20$ fps, or $79.0\%$ at $62$ fps as a performance/speed tradeoff.
\footnote{Code is available at \url{http://mmlab.ie.cuhk.edu.hk/projects/ST-Lattice/}}

\end{abstract}

%%%%%%%%% BODY TEXT

% !TEX root = ../main.tex

\section{Introduction}
\label{sec:intro}

%% Problem background

Object detection in videos has received increasing attention
as it sees immense potential in real-world applications
such as video-based surveillance.
Despite the remarkable progress in image-based
object detectors~\cite{girshick2015fast,ren2015faster,dai2016r},
extending them to the video domain remains challenging.
Conventional CNN-based methods~\cite{kang2016object,kang2017object}
typically detect objects on a per-frame basis and integrate
the results via temporal association and box-level post-processing.
Such methods are slow, resource-demanding, and often unable
to meet the requirements in real-time systems.
For example, a competitive detector based on Faster R-CNN~\cite{ren2015faster} can only
operate at $7$ \textit{fps} on a high-end GPU like Titan X.

%% Limitations of existing methods

A typical approach to this problem is to optimize the underlying networks,
\eg~via model compression~\cite{iandola2016squeezenet,howard2017mobilenets,zhang2017shufflenet}.
This way requires tremendous engineering efforts.
On the other hand, videos, by their special nature, provide
a different dimension for optimizing the detection framework.
Specifically, there exists strong continuity among
consecutive frames in a natural video, which suggests an alternative
way to reduce computational cost, that is, to propagate the computation
temporally.
Recently, several attempts along this direction were made,
\eg~tracking bounding boxes~\cite{kang2017object} or warping
features~\cite{zhu2017deep}. However, the improvement on the
overall performance/cost tradeoff remains limited --
the pursuit of one side often causes significant expense to the other.

\begin{figure}
    \centering
    \includegraphics[width=0.9\linewidth]{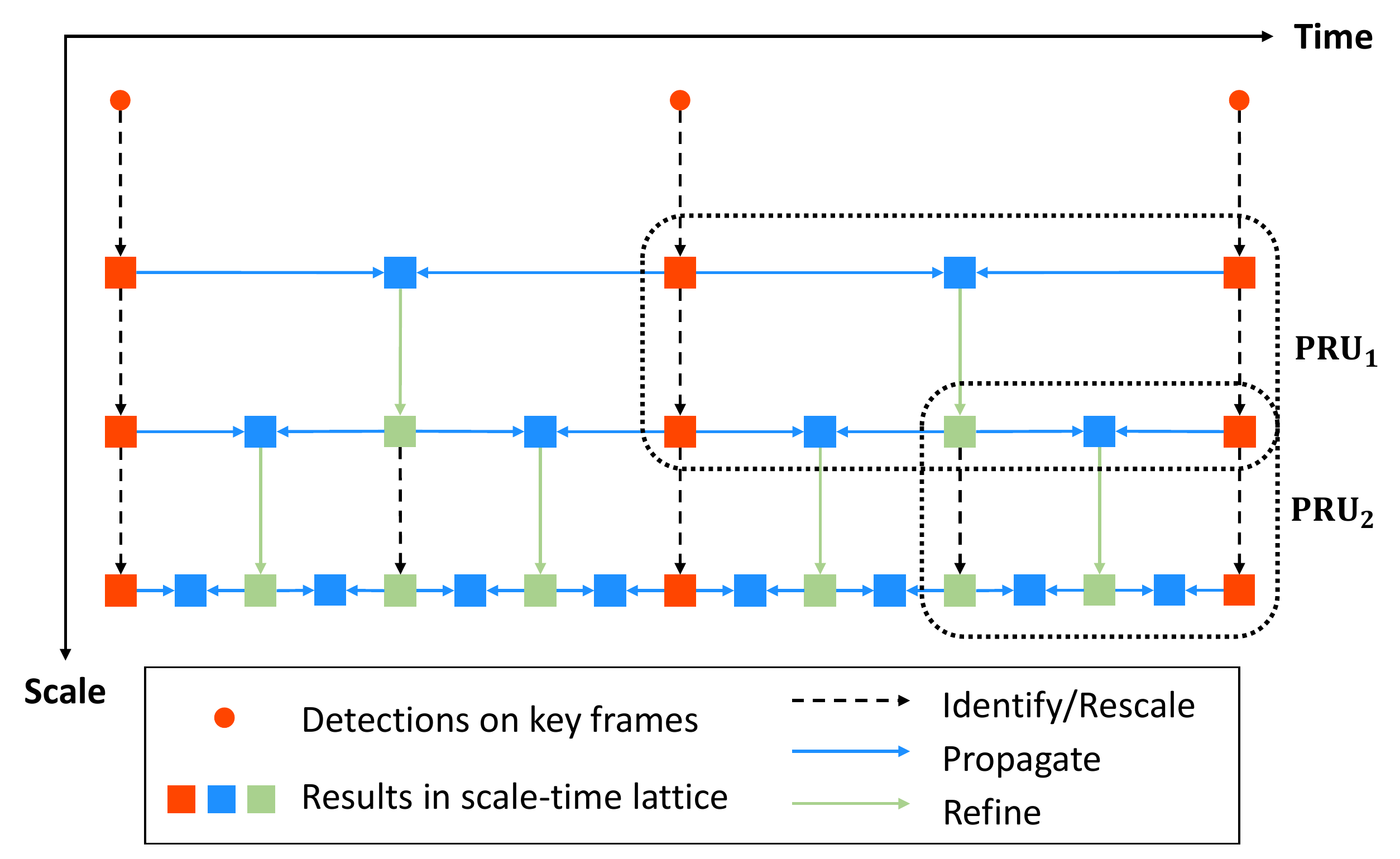}
    \caption{The proposed Scale-Time Lattice permits a flexible design space for performance-cost tradeoff. %The figure shows the computation of an instantiation of plausible networks on this space.
    }
    \label{fig:teaser}
    \vspace{-0.2cm}
\end{figure}

%% Raise key questions

Moving beyond such limitations requires a \emph{joint} perspective.
Generally, detecting objects in a video is a multi-step process.
The tasks studied in previous work,
\eg~image-based detection, temporal propagation,
and coarse-to-fine refinement, are just individual steps
in this process.
While improvements on individual steps have been studied a lot,
a key question is still left open:
\emph{``what is the most cost-effective strategy to combine them?''}

%% High-level view: Scale-Time Lattice

Driven by this joint perspective, we propose to explore a new strategy,
namely pursuing a \emph{balanced} design over a \emph{Scale-Time Lattice},
as shown in Figure~\ref{fig:teaser}.
The Scale-Time Lattice is a unified formulation, where the steps
mentioned above are directed links between the nodes at different scale-time
positions.
From this unified view, one can readily see how different steps contribute and
how the computational cost is distributed.

%% The proposed design

More importantly, this formulation comes with a \emph{rich design space},
where one can flexibly reallocate computation on demand.
In this work, we develop a balanced design by leveraging this flexibility.
Given a video,
the proposed framework first applies expensive object detectors to
the key frames selected \emph{sparsely} and \emph{adaptively}
based on the object motion and scale, to obtain effective bounding boxes for propagation.
These boxes are then propagated to intermediate frames
and refined across scales (from coarse to fine), via substantially cheaper
networks.
For this purpose, we devise a new component based on motion history
that can propagate bounding boxes effectively and efficiently.
This framework remarkably reduces the amortized cost by
only invoking the detector sparsely, while maintaining competitive performance
with a cheap but very effective propagation component.
This also makes it convenient to seek a good performance/cost tradeoff,
\eg~by tuning the key frame selection strategy or
the network complexity at individual steps.

%% Highlight contribution & experimental results

The main contributions of this work lie in several aspects:
(1) the \emph{Scale-Time Lattice} that provides a joint perspective
and a rich design space,
(2) a detection framework devised thereon that achieves better
speed/accuracy tradeoff, and
(3) several new technical components, \eg~a network for more
effective temporal propagation and an adaptive scheme for keyframe selection.
Without bells-and-whistles, \eg~model ensembling and multi-scale testing,
we obtain competitive performance
on par with the method~\cite{zhu2017deep,zhu2017flow} that won ImageNet
VID challenges 2017, but with a significantly faster running speed
of 20 fps.

% !TEX root = ../main.tex

\section{Related Work}
\label{sec:related}

\noindent
\textbf{Object detection in images.}
Contemporary object detection methods have been dominated by deep CNNs,
most of which follow two paradigms, \emph{two-stage} and \emph{single-stage}.
A two-stage pipeline firstly generates region proposals,
which are then classified and refined.
In the seminal work~\cite{girshick2014rich}, Girshick \etal proposed R-CNN,
an initial instantiation of the two-stage paradigm.
More efficient frameworks have been developed since then.
Fast R-CNN~\cite{girshick2015fast} accelerates feature extraction by
sharing computation.
Faster R-CNN~\cite{ren2015faster} takes a step further by introducing
a Region Proposal Network (RPN) to generate region proposals, and sharing
features across stages.
Recently, new variants,
\eg~R-FCN~\cite{dai2016r}, FPN~\cite{lin2017feature},
and Mask R-CNN~\cite{he2017mask}, further improve the performance.
Compared to two-stage pipelines,
a single-stage method is often more efficient but less accurate.
Liu \etal~\cite{liu2016ssd} proposed Single Shot Detector (SSD),
an early attempt of this paradigm.
It generates outputs from default boxes on a pyramid of feature maps.
Shen \etal~\cite{shen2017dsod} proposed DSOD,
which is similar but based on DenseNet~\cite{huang2017densely}.
YOLO~\cite{redmon2016you} and YOLOv2~\cite{redmon2017yolo9000} present
an alternative that frames detection as a regression problem.
Lin \etal~\cite{lin2017focal} proposed the use of focal loss
along with RetinaNet, which tackles the imbalance between foreground
and background classes.

\vspace{5pt} \noindent
\textbf{Object detection in videos.}
Compared with object detection in images, video object detection was less
studied until the new VID challenge was introduced to ImageNet.
Han \etal~\cite{han2016seq} proposed Seq-NMS that builds high-confidence
bounding box sequences and rescores boxes to the average or maximum confidence.
The method serves as a post-processing step,
thus requiring extra runtime over per-frame detection.
Kang \etal~\cite{kang2016object,kang2017object} proposed a framework
that integrates per-frame proposal generation, bounding box tracking and
tubelet re-scoring.
It is very expensive, as it requires per-frame feature
computation by deep networks.
Zhu \etal~\cite{zhu2017deep} proposed an efficient framework that runs
expensive CNNs on sparse and regularly selected key frames.
Features are propagated to other frames with optical flow.
The method achieves 10$\times$ speedup than per-frame detection at the cost of
$4.4\%$ mAP drop (from $73.9\%$ to $69.5\%$).
Our work differs to~\cite{zhu2017deep} in that
we select key frames adaptively rather than at a fixed interval basis.
In addition, we perform temporal propagation in a scale-time lattice space
rather than once as in~\cite{zhu2017deep}.
Based on the aforementioned work, Zhu \etal~\cite{zhu2017flow} proposed to
aggregate nearby features along the motion path, improving the feature quality.
However, this method runs slowly at around $1$ fps
due to dense detection and flow computation.
Feichtenhofer \etal~\cite{feichtenhofer2017detect} proposed to
learn object detection and cross-frame tracking with a multi-task objective,
and link frame-level detections to tubes.
They do not explore temporal propagation, only perform interpolation
between frames.
There are also weakly supervised methods~\cite{misra2015watch,prest2012learning,chen2017discover} that learn object detectors from videos.

\vspace{5pt} \noindent
\textbf{Coarse-to-fine approaches.}
The coarse-to-fine design has been adopted for various problems such as
face alignment~\cite{zhang2014coarse,zhu2015face},
optical flow estimation~\cite{hu2016efficient,ilg2017flownet},
semantic segmentation~\cite{li2017not},
and super-resolution~\cite{huang2015single,lai2017deep}.
These approaches mainly adopt cascaded structures to refine results
from low resolution to high resolution.
Our approach, however, adopts the coarse-to-fine behavior in two dimensions,
both spatially and temporally.
The refinement process forms a 2-D Scale-Time Lattice space
that allows gradual discovery of denser and more precise bounding boxes.
%

% !TEX root = ../main.tex

\section{Scale-Time Lattice}
\label{sec:framework}

\begin{figure*}
	\centering
	\includegraphics[height=158pt]{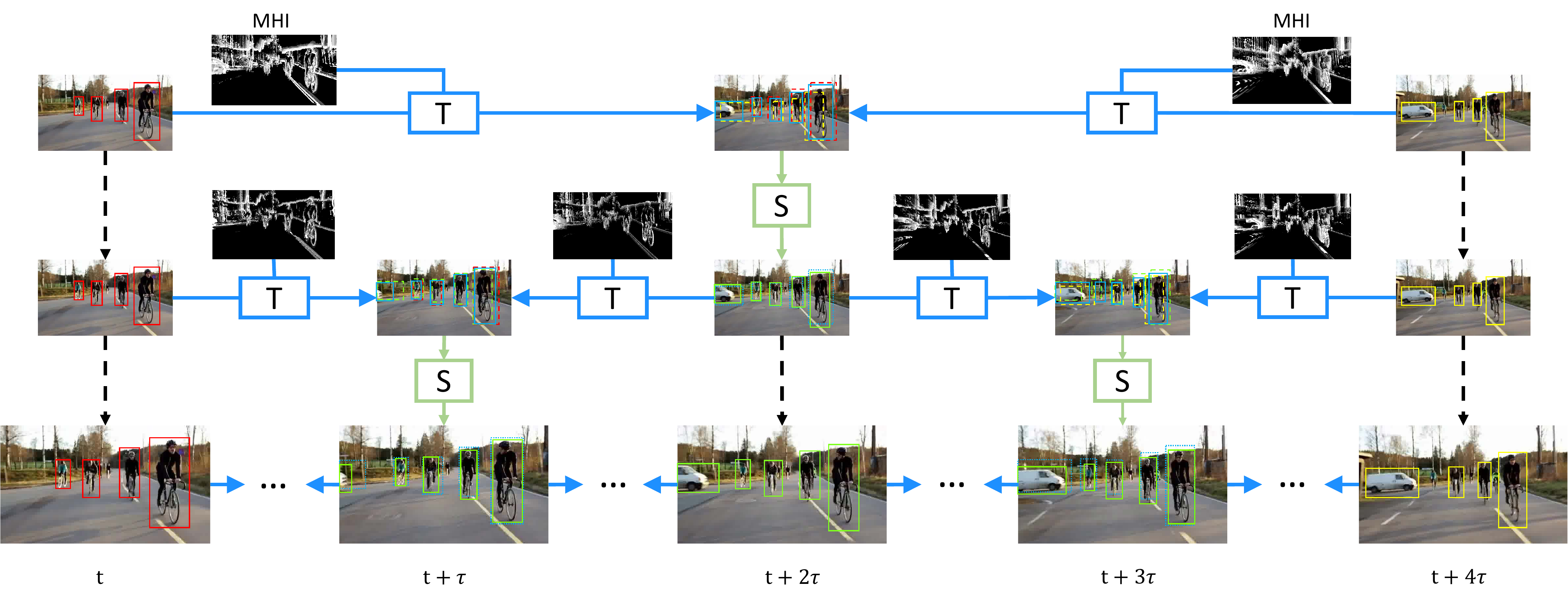}
	\vspace{-0.1cm}
	\caption{\small
		The Scale-Time Lattice, where each node represents the detection
		results at a certain scale and time point, and each edge represents
		an operation from one node to another.
		In particular, the horizontal edges (in blue color) represent
		the temporal propagation from one time step to the next; while
		the vertical edges (in green color) represent the spatial refinement
		from low to high resolutions.
		Given a video, the image-based detection is only done at sparsely
		chosen key frames, and the results are propagated along a pre-defined
		path to the bottom row. The final results at the bottom cover all time points.
	}
	\label{fig:stlattice}
\end{figure*}

%% Motivation

In developing a framework for video object detection, our primary goal
is to \emph{precisely} localize objects in each frame,
while meeting runtime requirements, \eg~high detection speed.
One way to achieve this is to apply the expensive object detectors
on as few key frames as possible, and rely on the spatial and temporal
connections to generate detection results for the intermediate frames.
While this is a natural idea, finding an optimal design is non-trivial.
In this work, we propose the \emph{Scale-Time Lattice}, which unifies the
sparse image-based detection and the construction of dense video detection
results into a single framework. A good balance of computational cost and
detection performance can then be achieved by carefully allocating resources
to different components within this framework.

%% Formulation of ST-Lattice

The \emph{Scale-Time Lattice}, as shown in Fig.~\ref{fig:stlattice},
is formulated as a directed acyclic graph.
Each node in this graph stands for the intermediate detection results
at a certain spatial resolution and time point,
in the form of bounding boxes.
The nodes are arranged in a way similar to a lattice:
from left to right, they follow the temporal order,
while from top to bottom, their scales increase gradually.
An edge in the graph represents
a certain operation that takes the detection results from the head node
and produces the detection results at the tail node.
In this work, we define two key operations, \emph{temporal propagation}
and \emph{spatial refinement},
which respectively correspond to the horizontal and vertical edges in the graph.
Particularly, the \emph{temporal propagation} edges connect nodes at the same
spatial scale but adjacent time steps.
The \emph{spatial refinement} edges connect nodes at the same time step
but neighboring scales.
Along this graph, detection results will be propagated from one node to
another via the operations introduced above following certain paths.
Eventually, the video detection results can be derived from the nodes
at the bottom row, which are at the finest scale and cover every time step.

%% Describe from a user view

On top of the Scale-Time Lattice, a video detection pipeline involves three steps:
1) generating object detection results on sparse key frames;
2) planning the paths from image-based detection results (input nodes) to
the dense video detection results (output nodes);
3) propagating key frame detection results to the intermediate frames
and refine them across scales.
The detection accuracy of the approach is measured at the output nodes.

%% Summary of benefits

The Scale-Time Lattice framework provides a rich design space for optimizing
the detection pipeline.
Since the total computational cost equals to the summation of
the cost on all paths, including the cost for invoking image-based detectors, it
is now convenient to seek a cost/performance tradeoff by well allocating the
budget of computation to different elements in the lattice.
For example, by sampling more key frames, we can improve
detection performance, but also introduce heavy computational cost.
On the other hand, we find that with much cheaper networks,
the propagation/refinement edges can carry the detection results over
a long path while still maintaining competitive accuracy.
Hence, we may obtain a much better accuracy/cost tradeoff
if the cost budget is used instead for the right component.
%reinforcing the temporal propagation and spatial refinement operations.

%% Distinctions

Unlike previous pursuits of accuracy/cost balance like spatial
pyramid or feature flow, the Scale-Time Lattice operates
from coarse to fine, both temporally and spatially.
The operation flow across the scale-time lattice narrows the temporal interval
while increasing the spatial resolution.
In the following section, we will describe the technical details of
individual operations along the lattice.

% !TEX root = ../main.tex

\section{Technical Design}
\label{sec:technical-design}

In this section, we introduce the design of key components in the
Scale-Time Lattice framework and show how they work together to
achieve an improved balance between performance and cost.
As shown in Figure~\ref{fig:teaser}, the lattice is comprised of
compound structures that connect with each other repeatedly to
perform temporal propagation and spatial refinement.
We call them \emph{Propagation and Refinement Units (PRUs)}.
After selecting a small number of key frames and obtaining the detection results
thereon, we propagate the results across time and scales via PRUs until they
reach the output nodes.
Finally, the detection results at the output nodes are integrated into
spatio-temporal tubes, and we use a tube-level classifier to reinforce the results.

\subsection{Propagation and Refinement Unit (PRU)}
\label{subsec:pru}

\begin{figure}
	\centering
	\includegraphics[width=0.68\linewidth]{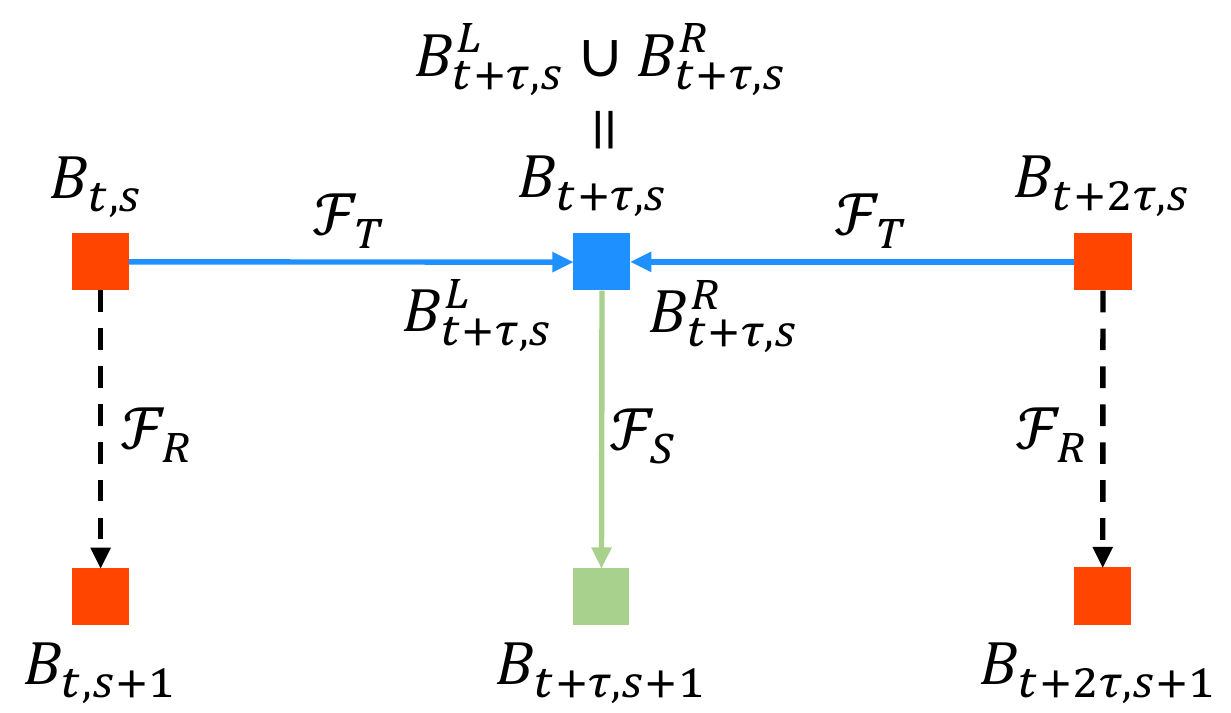}
	\caption{\small A Propagation and Refinement Unit.}
	\label{fig:pru}
	\vspace{-0.3cm}
\end{figure}

The PRU takes the detection results on two consecutive key frames as input,
propagates them to an intermediate frame,
and then refines the outputs to the next scale, as shown in Figure~\ref{fig:pru}.
Formally, we denote the detection results at time $t$ and scale level $s$
as $B_{t,s}=\{b_{t,s}^0,b_{t,s}^1,\dots,b_{t,s}^{n_t}\}$,
which is a set of bounding boxes $b_{t,s}^i=(x_{t,s}^i, y_{t,s}^i, w_{t,s}^i, h_{t,s}^i)$.
Similarly, we denote the ground truth bounding boxes as
$G_t=\{g_t^0,g_t^1,\dots,g_t^{m_t}\}$.
In addition, we use $I_t$ to denote the frame image at time $t$
and $M_{t \rightarrow t+\tau}$ to denote the motion representation from
frame $t$ to $t+\tau$.

A PRU at the $s$-level consists of a temporal propagation operator $\cF_T$,
a spatial refinement operator $\cF_S$, and a simple rescaling operator $\cF_R$.
Its workflow is to output $(B_{t,s+1}, B_{t+\tau,s+1}, B_{t+2\tau,s+1})$
given $B_{t,s}$ and $B_{t+2\tau,s}$.
The process can be formalized as
\begin{align}
	B_{t+\tau,s}^L &= \cF_T(B_{t,s}, M_{t\rightarrow t+\tau}), \\
	B_{t+\tau,s}^R &= \cF_T(B_{t+2\tau,s}, M_{t+2\tau\rightarrow t+\tau}), \\
	B_{t+\tau,s}   &= B_{t+\tau,s}^L \cup B_{t+\tau,s}^R, \\
	B_{t+\tau,s+1} &= \cF_S(B_{t+\tau,s}, I_{t+\tau}), \\
	B_{t,s+1}      &= \cF_R(B_{t, s}), \ B_{t+2\tau,s+1} = \cF_R(B_{t+2\tau, s}).
\end{align}
The procedure can be briefly explained as follows:
(1) $B_{t,s}$ is propagated temporally to the time step $t+\tau$ via $\cF_T$,
resulting in $B^L_{t+\tau,s}$.
(2) Similarly, $B_{t+2\tau,s}$ is propagated to the time step $t+\tau$
in an opposite direction, resulting in $B^R_{t+\tau,s}$.
(3) $B_{t+\tau,s}$, the results at time $t + \tau$, are then formed by their union.
(4) $B_{t+\tau,s}$ is refined to $B_{t+\tau,s+1}$ at the next scale via $\cF_S$.
(5) $B_{t,s+1}$ and $B_{t+2\tau,s+1}$ are simply obtained by rescaling
$B_{t, s}$ and $B_{t+2\tau, s}$.

Designing an effective pipeline of PRU is non-trivial.
Since the key frames are sampled sparsely to achieve high efficiency,
there can be large motion displacement and scale variance in between.
Our solution, as outlined above, is to factorize the workflow into two
key operations $\cF_T$ and $\cF_S$.
In particular, $\cF_T$ is to deal with the large motion displacement
between frames, taking into account the motion information.
This operation would roughly localize the objects at time $t+\tau$.
However, $\cF_T$ focuses on the object movement and it does not consider the
offset between the detection results $B_{t,s}$ and ground truth $G_t$.
Such deviation will be accumulated into the gap between
$B_{t+\tau,s}$ and $G_{t+\tau}$.
$\cF_S$ is designed to remedy this effect by regressing the bounding box
offsets in a coarse-to-fine manner, thus leading to more precise localization.
These two operations work together and are conducted iteratively following
the scale-time lattice to achieve the final detection results.

\vspace{-12pt}
\paragraph{Temporal propagation}
The idea of temporal propagation is previously explored in the video object
detection literatures~\cite{zhu2017deep,zhu2017flow,kang2016object}. Many of
these methods~\cite{zhu2017deep,zhu2017flow} rely on optical flow to
propagate detection results. In spite of its performance,
the approach is expensive for a real-time system and not tailored to
encoding the motion information over a long time span.
In our work, we adopt \emph{Motion History Image (MHI)}~\cite{bobick2001recognition}
as the motion representation which can be computed very efficiently and
preserve sufficient motion information for the propagation.

We represent the motion from time $t$ to $t + \tau$
as $M_{t\rightarrow t+\tau}=(H_{t\rightarrow t+\tau}, I_t^{(g)}, I_{t+\tau}^{(g)})$.
Here $H_{t\rightarrow t+\tau}$ denotes the MHI from $t$ to $t+\tau$,
and $I_t^{(g)}$ and $I_{t+\tau}^{(g)}$ denote the gray-scale images of the two frames
respectively, which serve as additional channels to enhance the motion expression with more details.
We use a small network (ResNet-18 in our experiments) with RoIAlign layer~\cite{he2017mask}
to extract the features of each box region. On top of the RoI-wise features,
a regressor is learned to predict the object movement from $t$ to $t+\tau$.

To train the regressor, we adopt a similar supervision to~\cite{kang2017object},
which learns the relative movement from $G_t$ to $G_{t+\tau}$.
The regression target of the $j$-th bounding box $\Delta_{\cF_T}^{*j}$
is defined as the relative movement between best overlapped ground truth box $g_t^j$
and the corresponding one on frame $g_{t+\tau}^j$,
adopting the same transformation and normalization used in most detection methods~\cite{girshick2014rich,girshick2015fast}.

\vspace{-12pt}
\paragraph{Coarse-to-fine refinement}
After propagation, $B_{t+\tau,s}$ is supposed to be around the
target objects but may not be precisely localized.
The refinement operator $\cF_S$ adopts a similar structure as the propagation operator
and aims to refine the propagated results.
It takes $I_{t+\tau}$ and the propagated boxes $B_{t+\tau,s}$ as
the inputs and yields refined boxes $B_{t+\tau,s+1}$.
The regression target $\Delta_{\cF_S}^*$ is calculated as the offset of
$B_{t+\tau,s}$ \wrt~$G_{t+\tau}$.
In the scale-time lattice,
smaller scales are used in early stages and larger scales are used in later stages.
Thereby, the detection results are refined in a coarse to fine manner.

\vspace{-12pt}
\paragraph{Joint optimization}
The temporal propagation network $\cF_T$ and spatial refinement network $\cF_S$ are
jointly optimized with a multi-task loss in an end-to-end fashion.
\begin{equation}
    \begin{split}
        &L(\Delta_{\cF_T}, \Delta_{\cF_S}, \Delta_{\cF_T}^*, \Delta_{\cF_S}^*) = \\
        &\frac{1}{N}\sum_{j=1}^N L_{\cF_T}(\Delta_{\cF_T}^j, \Delta_{\cF_T}^{*j}) 
        + \lambda\frac{1}{N}\sum_{j=1}^NL_{\cF_S}(\Delta_{\cF_S}^j, \Delta_{\cF_S}^{*j}),
    \end{split}
\end{equation}
where $N$ is the number of bounding boxes in a mini batch,
$\Delta_{\cF_T}$ and $\Delta_{\cF_S}$ are the network output of $\cF_T$ and $\cF_S$,
and $L_{\cF_T}$ and $L_{\cF_S}$ are smooth L1 loss of temporal propagation and spatial refinement network, respectively.

\subsection{Key Frame Selection and Path Planning}
\label{subsec:keyframe-selection}

Under the Scale-Time Lattice framework, selected key frames form the input
nodes, whose number and quality are critical to both detection accuracy and
efficiency. The most straightforward approach to key frame selection is uniform
sampling which is widely adopted in the previous methods
~\cite{zhu2017deep,feichtenhofer2017detect}.
While uniform sampling strategy is simple and effective, it ignores a key fact
that not all frames are equally important and effective for detection and propagation.
Thus a non-uniform frame selection strategy could be more desirable.

To this end, we propose an adaptive selection scheme based on our observation
that temporal propagation results tend to be inferior to single frame
image-based detection when the objects are small and moving quickly.
Thus the density of key frames should depend on propagation difficulty,
namely, we should select key frames more frequently in the presence
of small or fast moving objects.

The adaptive frame selection process works as follows.
We first run the detector on very sparse frames $\{t_0, t_1, t_2, \dots\}$
which are uniformly distributed.
Given the detection results, we evaluate how \emph{easy}
the results can be propagated, based on both the object size and motion.
The \emph{easiness measure} is computed as

\begin{align}
	e_{i,i+1} = \frac{1}{|I|}\sum_{(j,k)\in I}s_{t_i,t_{i+1}}^{j,k}m_{t_i,t_{i+1}}^{j,k}
\end{align}

where $I$ is the set of matched indices of $B^\prime_{t_i}$ and
$B^\prime_{t_{i+1}}$ through bipartite matching based on confidence score and bounding box IoUs,
$s_{t_i,t_{i+1}}^{j,k}=\frac{1}{2}(\sqrt{\text{area}(b_{t_i}^j)} + \sqrt{\text{area}(b_{t_{i+1}}^k)})$ is the object size measure
and $m_{t_i,t_{i+1}}^{j,k}=\text{IoU}(b_{t_i}^j, b_{t_{i+1}}^k)$ is the motion measure.
Note since the results can be noisy, we only consider boxes with high confidence
scores.
If $e_{i,i+1}$ falls below a certain threshold, an extra key frame
$\bar{t}_{i,i+1}=\frac{t_i+t_{i+1}}{2}$ is added.
This process is conducted for only one pass in our experiments.

With the selected key frames, we propose a simple scheme to plan the paths
in the scale-time lattice from input nodes to output nodes.
In each stage, we use propagation edges to link the nodes at the different time steps,
and then use a refinement edge to connect the nodes across scales.
Specifically, for nodes $(t_i, s)$ and $(t_{i+1}, s)$ at time point $t_i$ and $t_{i+1}$
of the scale level $s$, results are propagated to $((t_i + t_{i+1})/2, s)$,
then refined to $((t_i + t_{i+1})/2, s+1)$.
We set the max number of stages to $2$.
After two stages, we use linear interpolation as a very cheap propagation approach
to generate results for the remaining nodes.
More complex path planning may further improve the performance at the same cost,
which is left for future work.

\subsection{Tube Rescoring}
\label{subsec:tube-rescoring}

By associating the bounding boxes across frames at the last stage with the
propagation relations, we can construct \emph{object tubes}.
Given a linked tube $\cT=(b_{t_0}, \dots, b_{t_n})$ consisting of $t_n-t_0$
bounding boxes that starts from frame $t_0$ and terminates at $t_n$
with label $l$ given by the original detector,
we train a R-CNN like classifier to re-classify it following the scheme of
Temporal Segment Network (TSN)~\cite{wang2016temporal}.
During inference, we uniformly sample $K$ cropped bounding boxes from each tube
as the input of the classifier, and the class scores are fused to yield a \emph{tube-level} prediction.
After the classification, scores of bounding boxes in $\cT$ are adjusted by the
following equation.
\[
    s_i =
    \begin{cases}
        s_i + s^\prime, & \text{if } l = l^\prime \\
        \frac{1}{n}\sum_{i=0}^n s_i, & \text{otherwise}
    \end{cases}
\]
where $s_i$ is the class score of $b_{t_i}$ given by the detector,
and $s^\prime$ and $l^\prime$ are the score and label prediction of $\cT$ given by the classifier.
After the rescoring, scores of hard positive samples can be boosted and
false positives are suppressed.

% !TEX root = ../main.tex

\section{Experiments}
\label{sec:experiments}

\subsection{Experimental Setting}
\label{subsec:experiment-setup}

\noindent\textbf{Datasets.}
We experiment on the ImageNet VID dataset\footnote{\url{http://www.image-net.org/challenges/LSVRC/}}, a large-scale
benchmark for video object detection, which contains 3862 training videos and
555 validation videos with annotated bounding boxes of 30 classes. Following the standard practice, we train our models on the training set and measure the performance on the validation set using the mean average precision (mAP) metric.
We use a subset of ImageNet DET dataset
and VID dataset to train our base detector, following~\cite{kang2016object,zhu2017flow,feichtenhofer2017detect}.

\noindent\textbf{Implementation details.}
We train a Faster R-CNN as our base detector. We use ResNet-101 as the backbone network and select 15 anchors corresponding to 5 scales and 3 aspect ratios for the RPN.
A total of 200k iterations of SGD training is performed on 8 GPUs.
We keep boxes with an objectness score higher than 0.001, which results in
a mAP of $74.5$ and a recall rate of 91.6 with an average of 37 boxes per image.
During the joint training of PRU, two random frames are sampled from a video with a
temporal interval between $6$ and $18$.
We use the results of the base detector as input ROIs for propagation.
To obtain the MHI between frame $t$ and $t+\tau$, we uniformly sample
five images apart from frame $t$ and $t+\tau$ when $\tau$ is larger than $6$ for
acceleration.
The batch size is set to 64 and each GPU holds 8 images in each iteration.
Training lasts 90 epochs with a learning rate of $0.002$ followed by 30 epochs
with a learning rate of $0.0002$.
At each stage of the inference, we apply non-maximum suppression (NMS) with a
threshold $0.5$ to bidirectionally propagated boxes with the same class label
before they are further refined.
The propagation source of suppressed boxes are considered as linked with that
of reserved ones to form an object tube.
For the tube rescoring, we train a classifier with the backbone of ResNet-101
and the $K=6$ frames are sampled from each tube during inference.

\subsection{Results}
\label{subsec:results}
\begin{figure}
    \centering
    \includegraphics[width=0.9\linewidth]{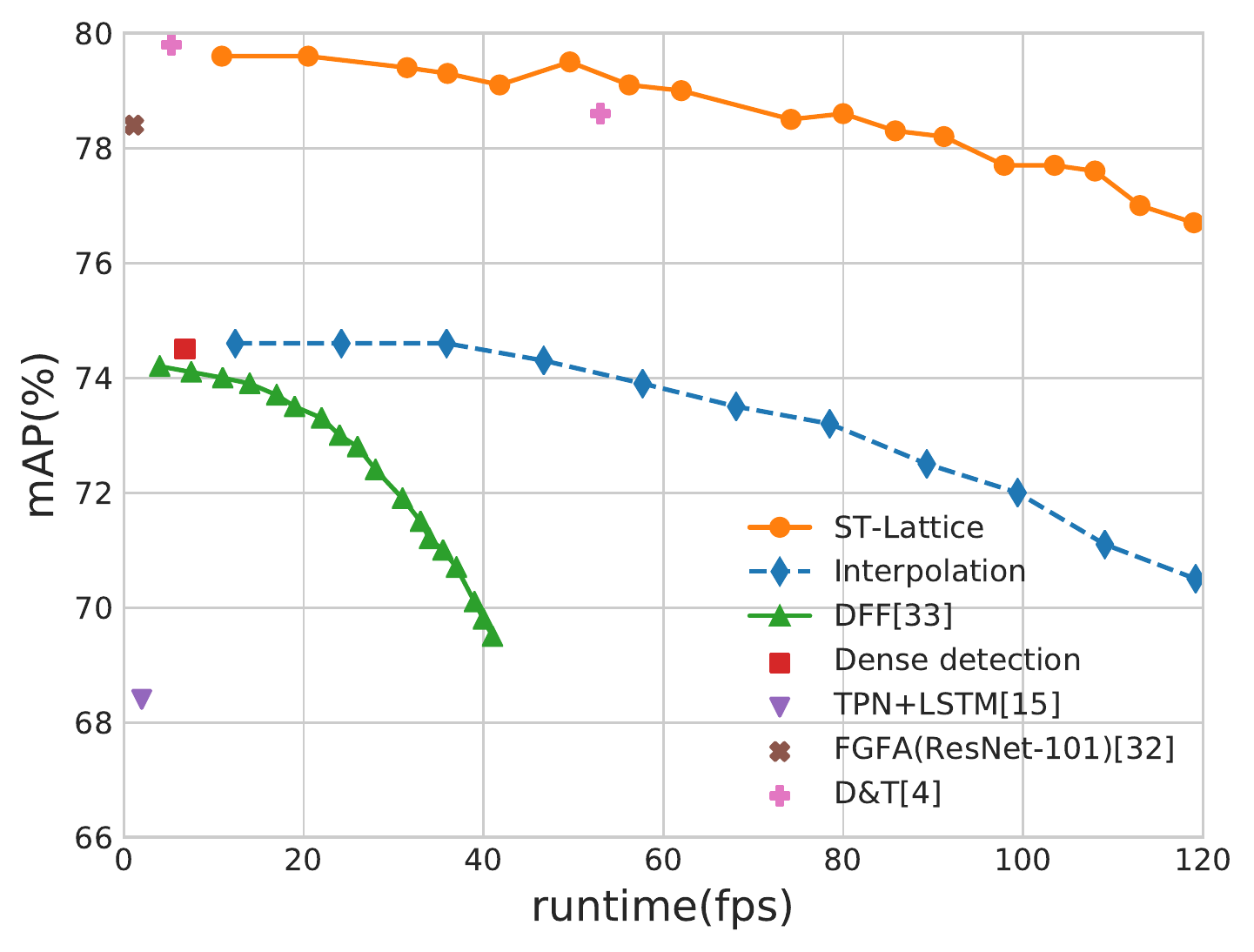}
    \vskip -0.45cm
    \caption[Performance and runtime on ImageNet VID dataset compared with existing methods.]{Performance and runtime on ImageNet VID dataset compared with existing methods.\footnotemark}
    \label{fig:overall-results}
    \vskip -0.2cm
\end{figure}

We summarize the cost/performance curve of our approach designed based on Scale-Time Lattice (ST-Lattice) and existing methods in Figure~\ref{fig:overall-results}.
The tradeoff is made under different temporal intervals.
The proposed ST-Lattice is clearly better than baselines such as na\"{i}ve interpolation and
DFF~\cite{zhu2017deep} which achieves a real-time detection rate by using optical flow to propagate features.
ST-Lattice also achieves better tradeoff than state-of-the-art methods, including D\&T~\cite{feichtenhofer2017detect}, TPN$+$LSTM~\cite{kang2017object}, and FGFA~\cite{zhu2017flow}.
In particular, our method achieves a mAP of $79.6$ at $20$ fps, which is competitive with D\&T\cite{feichtenhofer2017detect} that achieves $79.8$ at about $5$ fps. After a tradeoff on key frame selection, our approach still maintains a mAP of $79.0$ at an impressive $62$ fps.
We show the detailed class-wise performance in the supplementary material.

\begin{figure}
    \centering
    \includegraphics[width=0.9\linewidth]{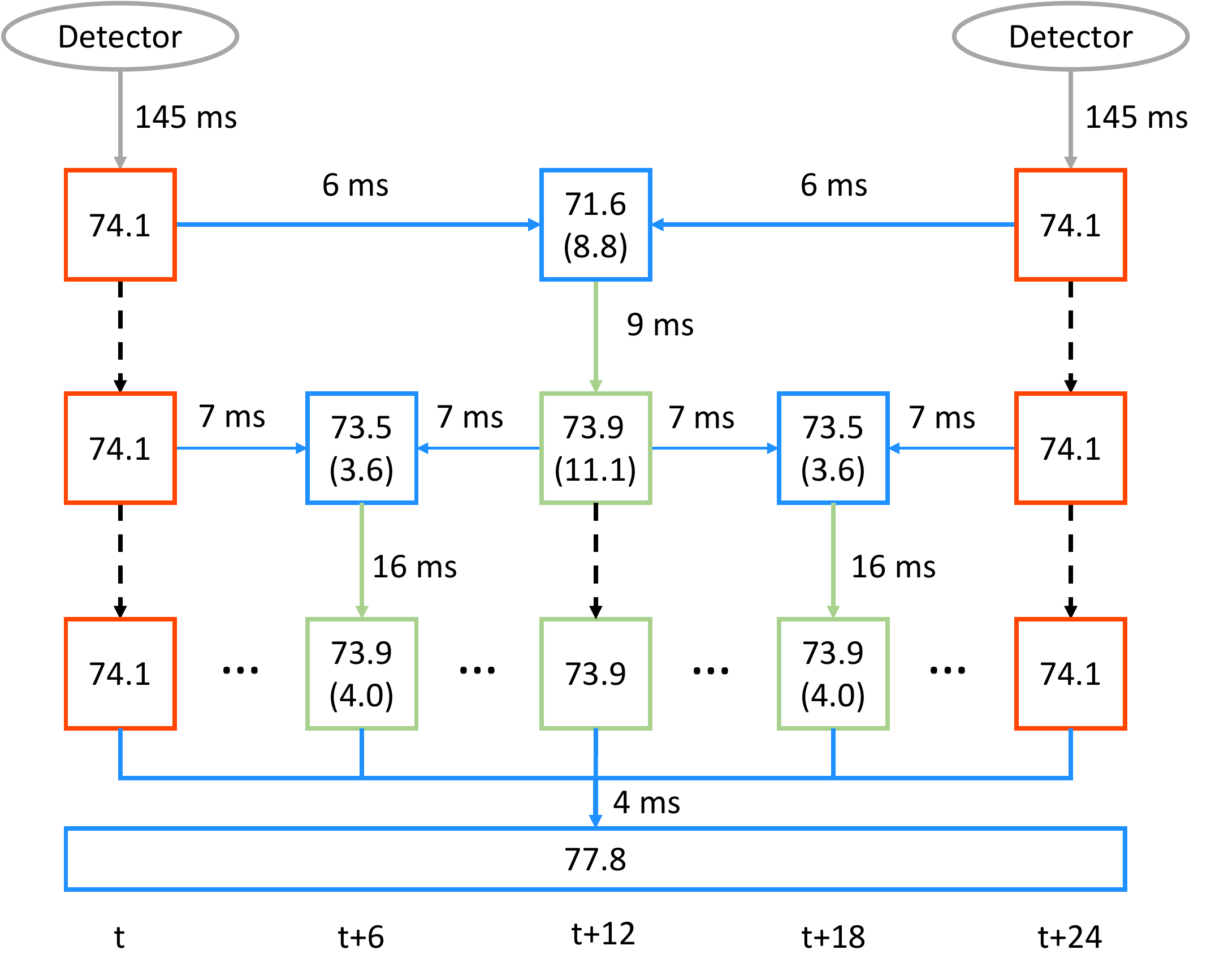}
    \vspace{-0.4cm}
    \caption{Cost allocation and mAP in the Scale-Time Lattice space.
        The value in the parenthesis refers to the improvement relative to interpolation.}
    \label{fig:cost-allocation}
    \vspace{-0.2cm}
\end{figure}

\begin{figure*}[hbt]
    \centering
    \includegraphics[width=0.88\linewidth]{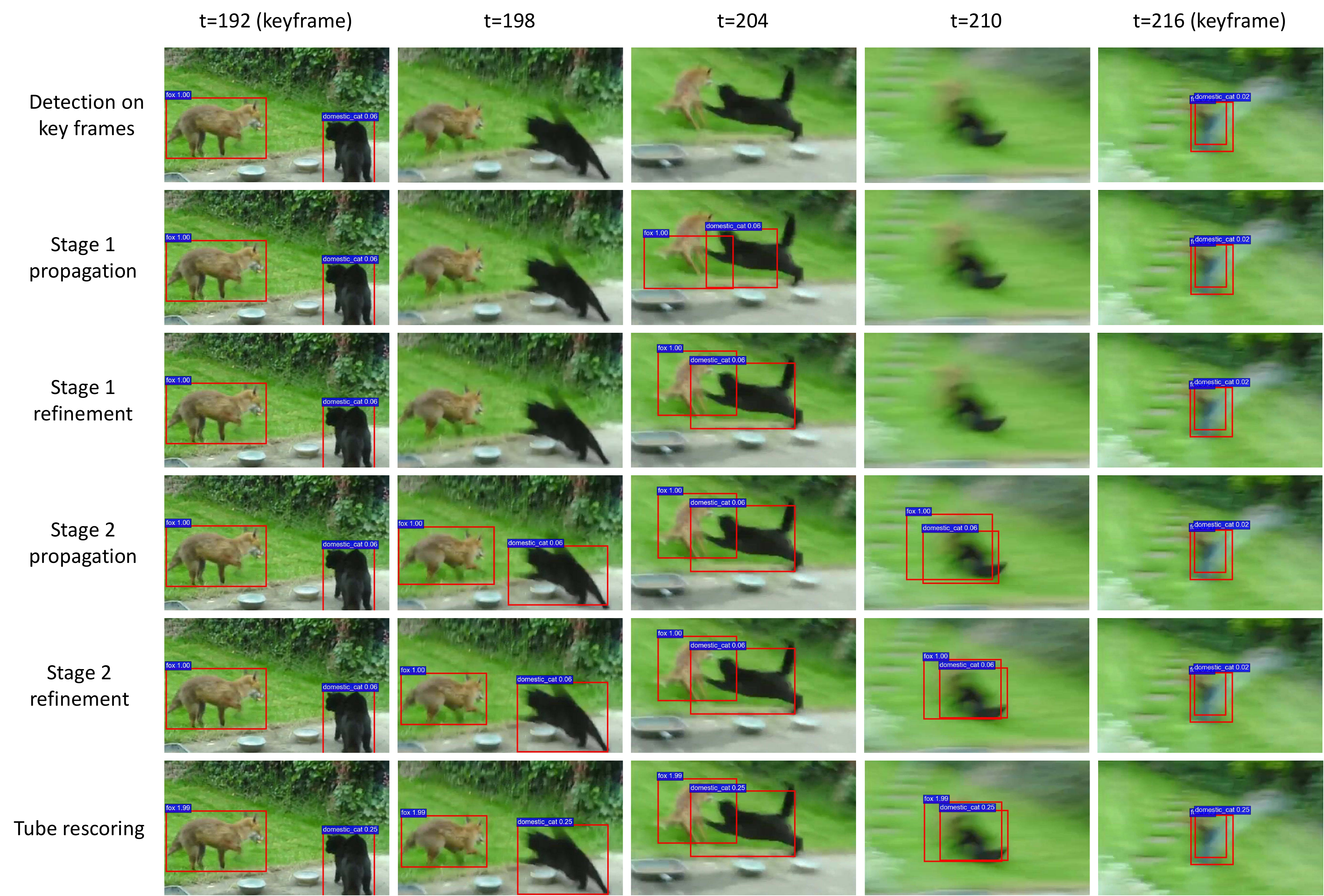}
    \caption{\small Examples video clips of the proposed Scale-Time Lattice.
        The per-frame baseline and detection results in different nodes
        are shown in different rows.}
    \label{fig:examples}
    \vspace{-0.2cm}
\end{figure*}

To further demonstrate how the performance and computational cost are balanced using the ST-Lattice space, we pick a configuration (with a fixed key frame interval of 24) and show the time cost of each edge and the mAP of each node in Figure~\ref{fig:cost-allocation}.
Thanks to the ST-Lattice, we can flexibly seek a suitable configuration to meet a variety of demands.
We provide some examples in Fig.~\ref{fig:examples}, showing the results of
per frame baseline and different nodes in the proposed ST-Lattice.

\footnotetext{The mAP is evaluated on all frames, except for the fast version of D\&T, which is evaluated on sparse key frames. We expect its performance will be lower in the full all-frame evaluation if the detections on other frames are interpolated.}

\subsection{Ablation Study}
\label{subsec:ablation-study}

In the following ablation study, we use a fixed key frame interval of 24 unless otherwise indicated
and run only the first stage of our approach.

\begin{figure}
    \centering
    \includegraphics[width=0.9\linewidth]{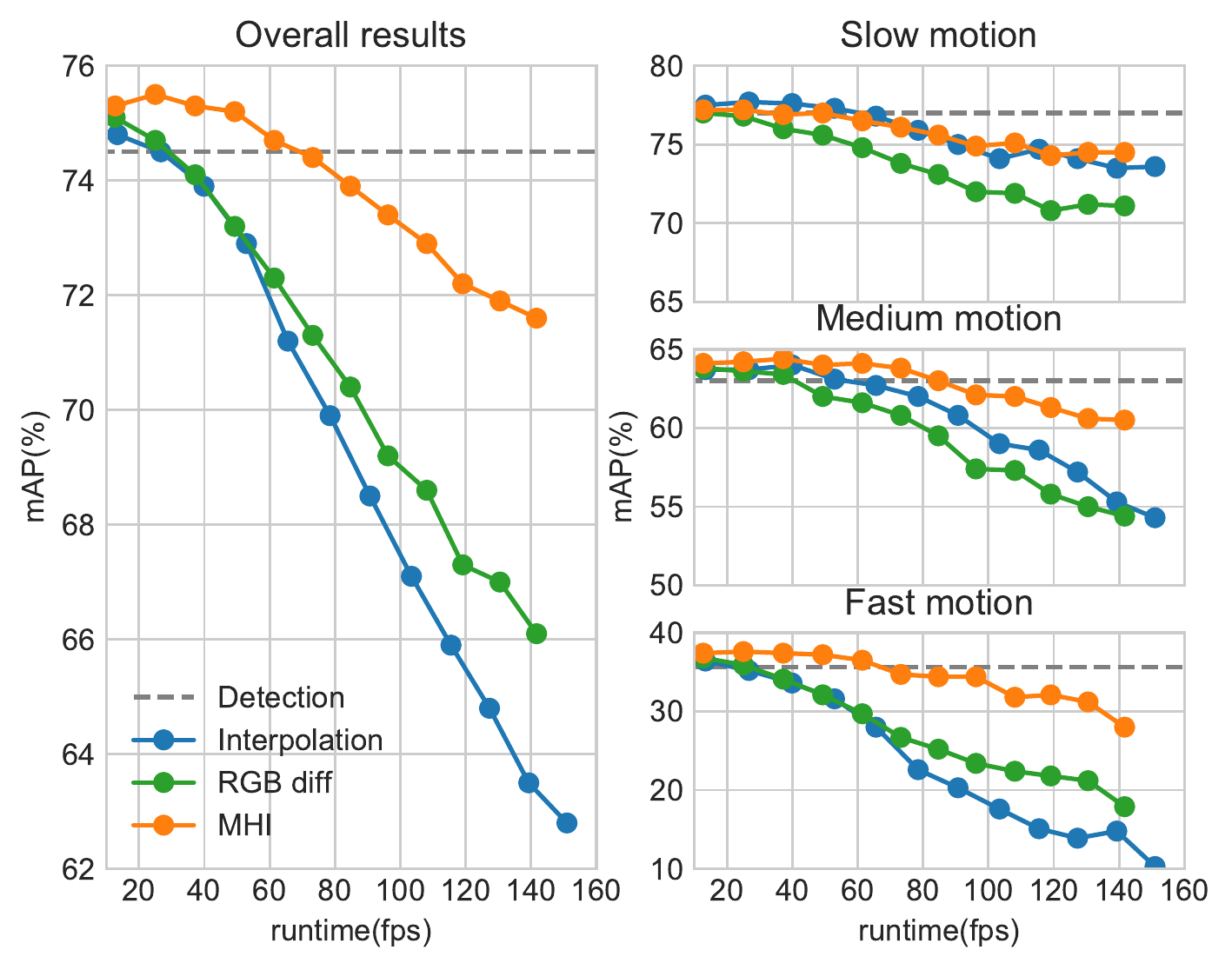}
    \vskip -0.3cm
    \caption{Results of different propagation methods under different key frame intervals. (Left) the overall results. (Right) Detailed results based on different object motion.}
    \label{fig:propagation}
    \vspace{-0.4cm}
\end{figure}

\vspace{0.1cm}
\noindent
\textbf{Temporal propagation.}
In the design space of ST-Lattice, there are many propagation methods
that can be explored. We compare the proposed propagation module with other
alternatives, such as linear interpolation and RGB difference based regression,
under different temporal intervals.
For a selected key frame interval, we evaluate the mAP of propagated results on the intermediate frame
from two consecutive key frames, without any refinement or rescoring.
We use different intervals (from 2 to 24) to see the balance between runtime and mAP.
Results are shown in Figure~\ref{fig:propagation}.
The fps is computed \wrt the detection time plus propagation/interpolation time.
The MHI based method outperforms other baselines by a large margin.
It even surpasses per-frame detection results when the temporal interval is small ($10$ frames apart). To take a deeper look into the differences of those propagation methods, we divide the ground truths into three parts according to object motion following~\cite{zhu2017flow}. We find that the gain mainly originates from objects with fast motion, which are considered more difficult than those with slow motion.
%We claim that MHI can capture the object motion across a long time span.

\begin{figure}
    \centering
	\includegraphics[width=0.9\linewidth]{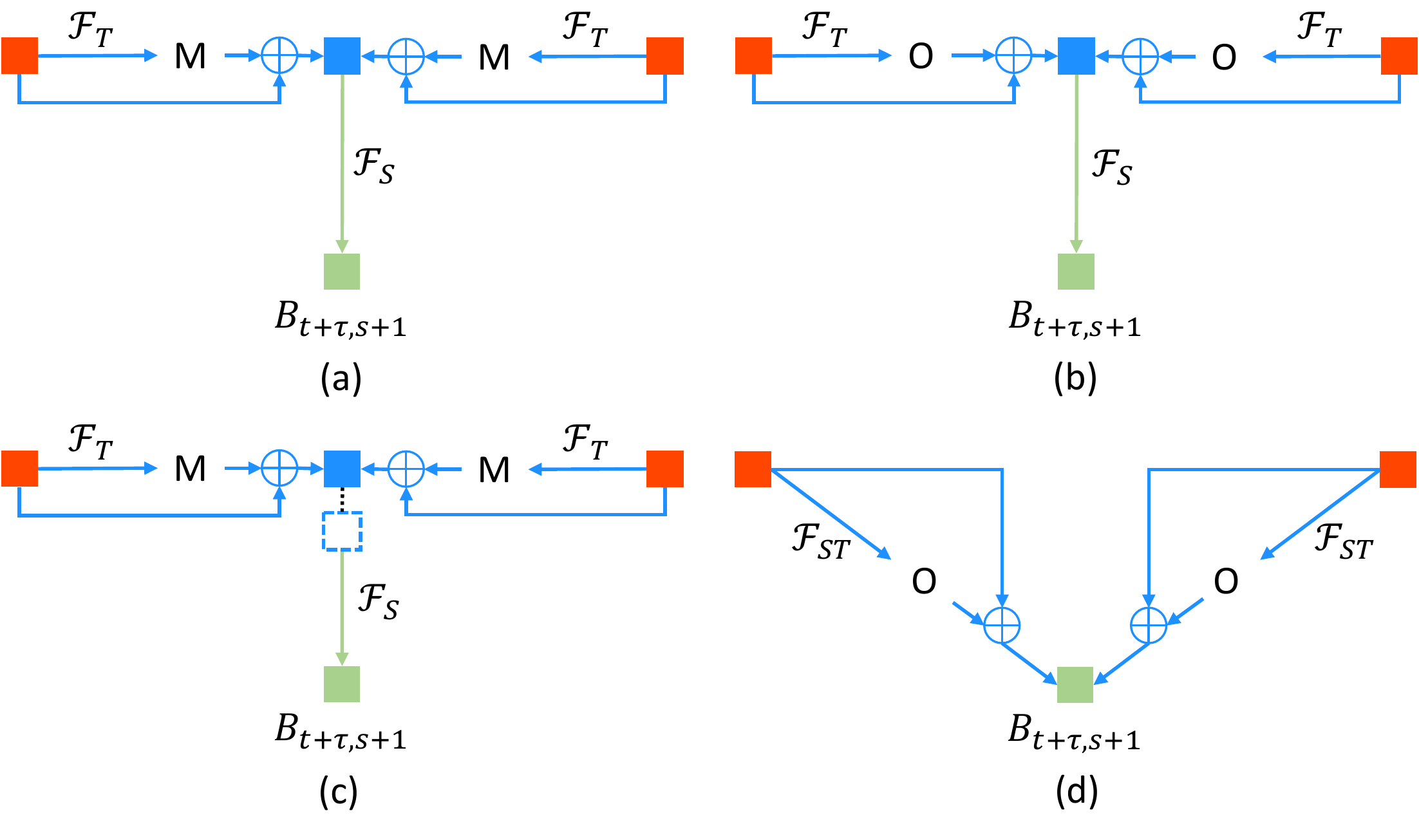}
	\vspace{-0.2cm}
    \caption{Variants of the basic unit.
        (a) is our design in Section~\ref{sec:technical-design} that regresses
        motion and offset respectively at two stages;
        (b) is a variant of our design that regresses the overall offset instead
        of motion at the first stage;
        (c) is the same of (a) in structure but not trained jointly;
        (d) is a single-step regression unit.}
    \label{fig:basic-unit-variants}
\end{figure}

\vspace{0.1cm}
\noindent
\textbf{Designs of PRU.}
Our design of the basic unit is a two-step regression component PRU that takes
the $B_{t,s}$ and $B_{t+2\tau,s}$ as input and outputs $B_{t+\tau,s+1}$.
Here, we test some variants of PRU as well as a single-step regression module,
as shown in Figure~\ref{fig:basic-unit-variants}.
$M$ represents motion displacement and $O$ denotes the offset \wrt the ground truth.
The results are shown in Table~\ref{tab:basic-unit}.
We find that design (a) that decouples the estimation of temporal motion displacement and spatial offset simplifies the learning target of regressors, thus yielding a better results than designs (b) and (d).
In addition, comparing (a) and (c), joint training of two-stage regression also improves the results by back propagating the gradient of the refinement component to the propagation component, which in turn increases the mAP of the first step results.

\begin{table}[htb]
    \centering
    \caption{Performance of different designs of basic unit. $v_T$ and $v_S$ refers to
        $B_{t+\tau,s}$ (the blue node) and $B_{t+\tau,s+1}$ (the green node) in Figure~\ref{fig:basic-unit-variants},  respectively.}
    \small
    \begin{tabular}{*{4}{c}}
        \toprule
                 & $v_T$ mAP (\%) & $v_S$ mAP (\%) & Runtime (ms) \\
        \midrule
        unit (a) & 71.6           & 73.9           & 21           \\
        unit (b) & 70.6           & 72.1           & 21           \\
        unit (c) & 71.4           & 73.7           & 21           \\
        unit (d) & N/A            & 71.0           & 12           \\
        \bottomrule
    \end{tabular}
    \label{tab:basic-unit}
\end{table}

\vspace{0.1cm}
\noindent
\textbf{Cost allocation.}
We investigate different cost allocation strategies by trying networks of
different depths for the propagation and refinement components.
Allocating computational costs at different edges on the ST-Lattice
would not have the same effects,
so we test different strategies by replacing the network of propagation
and refinement components with cheaper or more expensive ones.
The results in Table~\ref{tab:network-size} indicate that the performance
increases as the network gets deeper for both the propagation and refinement.
Notably, it is more fruitful to use a deeper network for the spatial refinement
network than the temporal propagation network.
Specifically, keeping the other one as medium, increasing the network size of spatial refinement from small to large results in a gain of $1.2$ mAP ($72.5\rightarrow73.7$),
while adding the same computational cost on $\cF_T$ only leads to an improvement of $0.8$ mAP ($72.7\rightarrow73.5$).

\begin{table}[t]
    \centering
    \caption{Performance of different combinations of propagation (T) and
        refinement (S) components. The two numbers ($x$/$y$) represent the mAP after propagation and after spatial refinement, respectively. \textit{Small}, \textit{medium} and \textit{large} refers to channel-reduced ResNet-18, ResNet-18 and ResNet-34.}
    \small
    \begin{tabular}{c|c|ccc}
        \hline
        \multicolumn{2}{l|}{\multirow{2}{*}{}} & \multicolumn{3}{c}{Net S}                                     \\
        \cline{3-5}
        \multicolumn{2}{l|}{}                  & small                     & medium    & large                 \\
        \hline
        \multirow{3}{*}{\rotatebox{90}{Net T}} & small                     & 67.7/71.1 & 67.7/72.7 & 67.8/72.6 \\
                                               & Medium                    & 71.5/72.5 & 71.6/73.9 & 71.5/73.7 \\
                                               & Large                     & 72.8/73.1 & 72.0/73.5 & 71.8/74.2 \\
        \hline
    \end{tabular}

    \label{tab:network-size}
\end{table}

\begin{figure}
    \centering
	\includegraphics[width=0.9\linewidth]{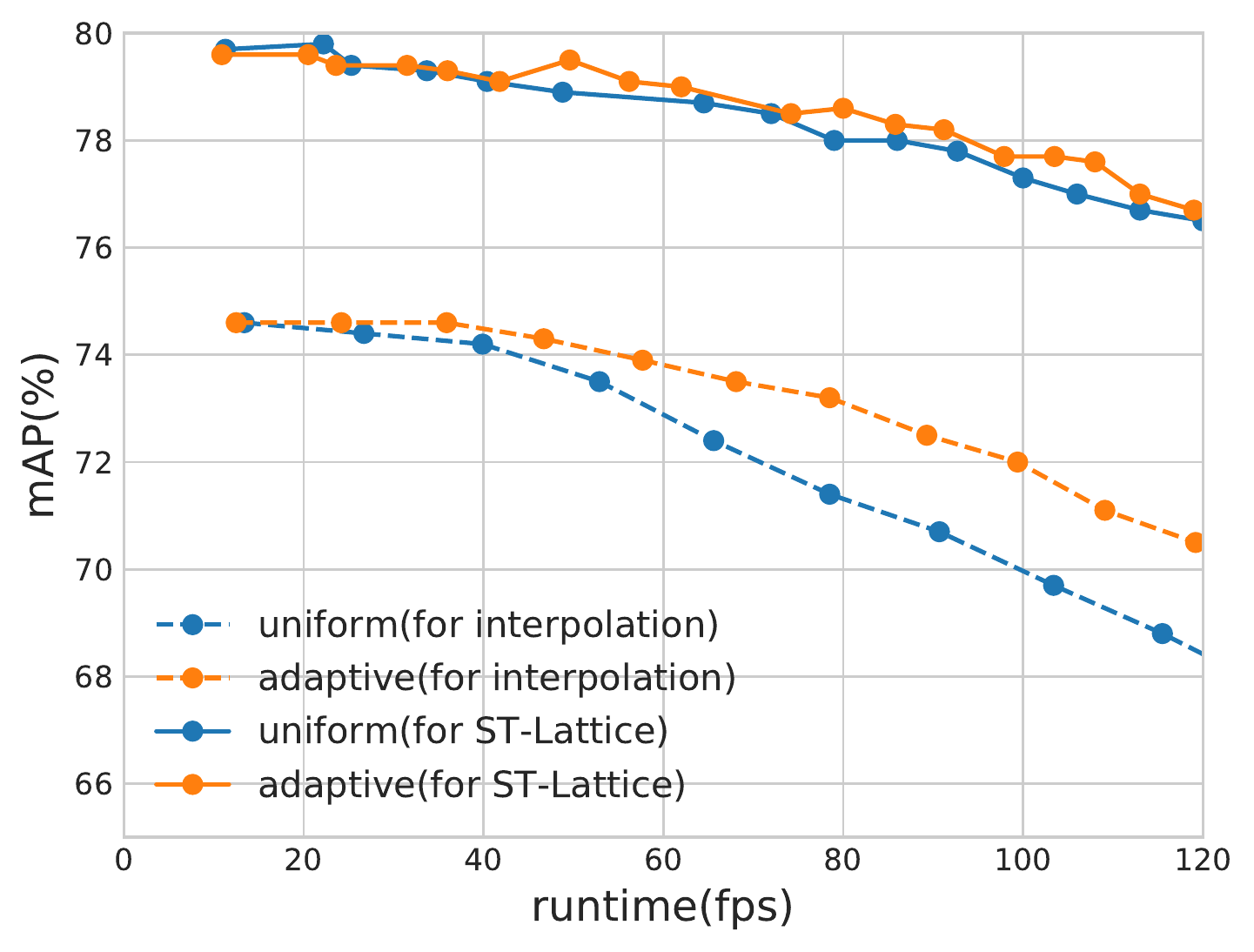}
	\vspace{-0.2cm}
    \caption{Uniformly sampled and adaptively selected key frames.}
	\label{fig:keyframe}
	\vspace{-0.3cm}
\end{figure}

\vspace{0.1cm}
\noindent
\textbf{Key frame selection.}
The selection of input nodes is another design option available on the ST-Lattice.
In order to compare the effects of different key frame selection strategies,
we evaluate the na\"{i}ve interpolation approach and the proposed ST-Lattice based on uniformly sampled and adaptively selected key frames.
The results are shown in Figure~\ref{fig:keyframe}.
% For both methods, adaptive key frame selection is superior to uniform sampling,
% and a larger performance gain is seen by using adaptive scheme if the key frames are distributed more sparsely.
For the na\"{i}ve interpolation, the adaptive scheme leads to a large performance gain.
Though the adaptive key frame selection does not bring as much improvement to ST-Lattice as interpolation, it is still superior to uniform sampling. Especially, its advantage stands out when the interval gets larger.
%leading to about $1\sim2$ mAP improvement with the same runtime, or a $30\%$ speedup under the same performance.
%
Adaptive selection works better because through our formulation, more hard samples are selected for running per-frame detector (rather than propagation) and leave easier samples for propagation.
This phenomenon can be observed when we quantify the mAP of detections on adaptively selected key frames than uniformly sampled ones ($73.3$ vs $74.1$), suggesting that more harder samples are selected by the adaptive scheme.

%To investigate how the adaptive selection works, we measure the mAP of detection results on key frames and find out that it is lower on adaptively selected key frames than uniformly sampled ones (73.3 vs 74.5), suggesting that more key frames are
%selected on hard samples and this distribution can reduce the error of the propagation.

% !TEX root = ../main.tex

\section{Conclusion}
\label{sec:conclusion}

We have presented the Scale-Time Lattice, a flexible framework that offers a rich design space to balance the performance and cost in video object detection.
It provides a joint perspective that integrates detection, temporal propagation, and across-scale refinement.
We have shown various configurations designed under this space and demonstrated their competitive performance against state-of-the-art video object detectors with much faster speed.
The proposed Scale-Time Lattice is not only useful for designing algorithms for video object detection, but also can be applied to other video-related domains such as video object segmentation and tracking.

\vspace{-0.2cm}
\paragraph{Acknowledgment}

This work is partially supported by the Big Data Collaboration Research grant from SenseTime Group (CUHK Agreement No. TS1610626), the Early Career Scheme (ECS) of Hong Kong (No. 24204215), and the General Research Fund (GRF) of Hong Kong (No. 14236516).

\clearpage
\clearpage

{\small
    \bibliographystyle{ieee}
    \bibliography{sections/long.bib,sections/egbib.bib}

\begin{thebibliography}{10}\itemsep=-1pt

\bibitem{bobick2001recognition}
A.~F. Bobick and J.~W. Davis.
\newblock The recognition of human movement using temporal templates.
\newblock {\em IEEE Transactions on Pattern Analysis and Machine Intelligence},
  23(3):257--267, 2001.

\bibitem{chen2017discover}
K.~Chen, H.~Song, C.~C. Loy, and D.~Lin.
\newblock Discover and learn new objects from documentaries.
\newblock In {\em IEEE Conference on Computer Vision and Pattern Recognition},
  2017.

\bibitem{dai2016r}
J.~Dai, Y.~Li, K.~He, and J.~Sun.
\newblock R-fcn: Object detection via region-based fully convolutional
  networks.
\newblock In {\em Advances in Neural Information Processing Systems}, 2016.

\bibitem{feichtenhofer2017detect}
C.~Feichtenhofer, A.~Pinz, and A.~Zisserman.
\newblock Detect to track and track to detect.
\newblock In {\em IEEE International Conference on Computer Vision}, 2017.

\bibitem{girshick2015fast}
R.~Girshick.
\newblock Fast r-cnn.
\newblock In {\em IEEE International Conference on Computer Vision}, 2015.

\bibitem{girshick2014rich}
R.~Girshick, J.~Donahue, T.~Darrell, and J.~Malik.
\newblock Rich feature hierarchies for accurate object detection and semantic
  segmentation.
\newblock In {\em IEEE Conference on Computer Vision and Pattern Recognition},
  2014.

\bibitem{han2016seq}
W.~Han, P.~Khorrami, T.~L. Paine, P.~Ramachandran, M.~Babaeizadeh, H.~Shi,
  J.~Li, S.~Yan, and T.~S. Huang.
\newblock Seq-nms for video object detection.
\newblock {\em arXiv preprint arXiv:1602.08465}, 2016.

\bibitem{he2017mask}
K.~He, G.~Gkioxari, P.~Doll\'{a}r, and R.~Girshick.
\newblock Mask r-cnn.
\newblock In {\em IEEE International Conference on Computer Vision}, 2017.

\bibitem{howard2017mobilenets}
A.~G. Howard, M.~Zhu, B.~Chen, D.~Kalenichenko, W.~Wang, T.~Weyand,
  M.~Andreetto, and H.~Adam.
\newblock Mobilenets: Efficient convolutional neural networks for mobile vision
  applications.
\newblock {\em arXiv preprint arXiv:1704.04861}, 2017.

\bibitem{hu2016efficient}
Y.~Hu, R.~Song, and Y.~Li.
\newblock Efficient coarse-to-fine patchmatch for large displacement optical
  flow.
\newblock In {\em IEEE Conference on Computer Vision and Pattern Recognition},
  2016.

\bibitem{huang2017densely}
G.~Huang, Z.~Liu, L.~van~der Maaten, and K.~Q. Weinberger.
\newblock Densely connected convolutional networks.
\newblock In {\em IEEE Conference on Computer Vision and Pattern Recognition},
  2017.

\bibitem{huang2015single}
J.-B. Huang, A.~Singh, and N.~Ahuja.
\newblock Single image super-resolution from transformed self-exemplars.
\newblock In {\em IEEE Conference on Computer Vision and Pattern Recognition},
  2015.

\bibitem{iandola2016squeezenet}
F.~N. Iandola, S.~Han, M.~W. Moskewicz, K.~Ashraf, W.~J. Dally, and K.~Keutzer.
\newblock Squeezenet: Alexnet-level accuracy with 50x fewer parameters and< 0.5
  mb model size.
\newblock {\em arXiv preprint arXiv:1602.07360}, 2016.

\bibitem{ilg2017flownet}
E.~Ilg, N.~Mayer, T.~Saikia, M.~Keuper, A.~Dosovitskiy, and T.~Brox.
\newblock Flownet 2.0: Evolution of optical flow estimation with deep networks.
\newblock In {\em IEEE Conference on Computer Vision and Pattern Recognition},
  2017.

\bibitem{kang2017object}
K.~Kang, H.~Li, T.~Xiao, W.~Ouyang, J.~Yan, X.~Liu, and X.~Wang.
\newblock Object detection in videos with tubelet proposal networks.
\newblock In {\em IEEE Conference on Computer Vision and Pattern Recognition},
  2017.

\bibitem{kang2016object}
K.~Kang, W.~Ouyang, H.~Li, and X.~Wang.
\newblock Object detection from video tubelets with convolutional neural
  networks.
\newblock In {\em IEEE Conference on Computer Vision and Pattern Recognition},
  2016.

\bibitem{lai2017deep}
W.-S. Lai, J.-B. Huang, N.~Ahuja, and M.-H. Yang.
\newblock Deep laplacian pyramid networks for fast and accurate
  super-resolution.
\newblock {\em arXiv preprint arXiv:1704.03915}, 2017.

\bibitem{li2017not}
X.~Li, Z.~Liu, P.~Luo, C.~C. Loy, and X.~Tang.
\newblock Not all pixels are equal: Difficulty-aware semantic segmentation via
  deep layer cascade.
\newblock In {\em IEEE Conference on Computer Vision and Pattern Recognition},
  2017.

\bibitem{lin2017feature}
T.-Y. Lin, P.~Dollar, R.~Girshick, K.~He, B.~Hariharan, and S.~Belongie.
\newblock Feature pyramid networks for object detection.
\newblock In {\em IEEE Conference on Computer Vision and Pattern Recognition},
  July 2017.

\bibitem{lin2017focal}
T.-Y. Lin, P.~Goyal, R.~Girshick, K.~He, and P.~Doll\'{a}r.
\newblock Focal loss for dense object detection.
\newblock In {\em IEEE International Conference on Computer Vision}, 2017.

\bibitem{liu2016ssd}
W.~Liu, D.~Anguelov, D.~Erhan, C.~Szegedy, S.~Reed, C.-Y. Fu, and A.~C. Berg.
\newblock Ssd: Single shot multibox detector.
\newblock In {\em European Conference on Computer Vision}, 2016.

\bibitem{misra2015watch}
I.~Misra, A.~Shrivastava, and M.~Hebert.
\newblock Watch and learn: Semi-supervised learning of object detectors from
  videos.
\newblock In {\em IEEE Conference on Computer Vision and Pattern Recognition},
  2015.

\bibitem{prest2012learning}
A.~Prest, C.~Leistner, J.~Civera, C.~Schmid, and V.~Ferrari.
\newblock Learning object class detectors from weakly annotated video.
\newblock In {\em IEEE Conference on Computer Vision and Pattern Recognition},
  2012.

\bibitem{redmon2016you}
J.~Redmon, S.~Divvala, R.~Girshick, and A.~Farhadi.
\newblock You only look once: Unified, real-time object detection.
\newblock In {\em IEEE Conference on Computer Vision and Pattern Recognition},
  2016.

\bibitem{redmon2017yolo9000}
J.~Redmon and A.~Farhadi.
\newblock Yolo9000: Better, faster, stronger.
\newblock In {\em IEEE Conference on Computer Vision and Pattern Recognition},
  2017.

\bibitem{ren2015faster}
S.~Ren, K.~He, R.~Girshick, and J.~Sun.
\newblock Faster r-cnn: Towards real-time object detection with region proposal
  networks.
\newblock In {\em Advances in Neural Information Processing Systems}, 2015.

\bibitem{shen2017dsod}
Z.~Shen, Z.~Liu, J.~Li, Y.-G. Jiang, Y.~Chen, and X.~Xue.
\newblock Dsod: Learning deeply supervised object detectors from scratch.
\newblock In {\em IEEE International Conference on Computer Vision}, 2017.

\bibitem{wang2016temporal}
L.~Wang, Y.~Xiong, Z.~Wang, Y.~Qiao, D.~Lin, X.~Tang, and L.~Van~Gool.
\newblock Temporal segment networks: Towards good practices for deep action
  recognition.
\newblock In {\em European Conference on Computer Vision}, 2016.

\bibitem{zhang2014coarse}
J.~Zhang, S.~Shan, M.~Kan, and X.~Chen.
\newblock Coarse-to-fine auto-encoder networks (cfan) for real-time face
  alignment.
\newblock In {\em European Conference on Computer Vision}, 2014.

\bibitem{zhang2017shufflenet}
X.~Zhang, X.~Zhou, M.~Lin, and J.~Sun.
\newblock Shufflenet: An extremely efficient convolutional neural network for
  mobile devices.
\newblock {\em arXiv preprint arXiv:1707.01083}, 2017.

\bibitem{zhu2015face}
S.~Zhu, C.~Li, C.~C. Loy, and X.~Tang.
\newblock Face alignment by coarse-to-fine shape searching.
\newblock In {\em IEEE Conference on Computer Vision and Pattern Recognition},
  2015.

\bibitem{zhu2017flow}
X.~Zhu, Y.~Wang, J.~Dai, L.~Yuan, and Y.~Wei.
\newblock Flow-guided feature aggregation for video object detection.
\newblock In {\em IEEE International Conference on Computer Vision}, 2017.

\bibitem{zhu2017deep}
X.~Zhu, Y.~Xiong, J.~Dai, L.~Yuan, and Y.~Wei.
\newblock Deep feature flow for video recognition.
\newblock In {\em IEEE Conference on Computer Vision and Pattern Recognition},
  2017.

\end{thebibliography}
}

\end{document}